\documentclass[runningheads]{llncs}
\usepackage[T1]{fontenc}
\usepackage{graphicx}
\usepackage{booktabs}
\usepackage{amssymb}
\usepackage{amsmath}
\usepackage{subcaption}
\begin{document}

\title{Bringing together invertible UNets with invertible attention modules for memory-efficient diffusion models}
\titlerunning{Bringing together invertible UNets with invertible attention modules}
%
%

\author{Karan Jain\orcidID{0009-0003-6435-3227} \and Mohammad Nayeem Teli\orcidID{0000-0002-8269-1479}}

\authorrunning{K. Jain and M. N. Teli}
%

\institute{University of Maryland, College Park MD 20742, USA \\
\email{\{kjain12,nayeem\}@umd.edu} }
\maketitle              
\begin{abstract}
Diffusion models have recently gained state of the art performance on many image generation tasks. However, most models require significant computational resources to achieve this. This becomes apparent in the application of medical image synthesis due to the 3D nature of medical datasets like CT-scans, MRIs, electron microscope, etc. In this paper we propose a novel architecture for a single GPU memory-efficient training for diffusion models for high dimensional medical datasets. The proposed model is built by using an invertible UNet architecture with invertible attention modules. This leads to the following two contributions: 1. denoising diffusion models and thus enabling memory usage to be independent of the dimensionality of the dataset, and 2. reducing the energy usage during training. While this new model can be applied to a multitude of image generation tasks, we showcase its memory-efficiency on the 3D BraTS2020 dataset leading to up to 15\% decrease in peak memory consumption during training with comparable results to SOTA while maintaining the image quality.

\end{abstract}
\section{Introduction}
\label{sec:1}

With the growing demands for GPU memory for generative models \cite{rajbhandari2021zeroinfinitybreakinggpumemory}, especially during training, it is becoming increasingly necessary to build memory-efficient systems. Additionally, energy concerns with generative models are rising as models tend to require multiple GPUs for training and inference. Training with multiple GPUs tends to lead to more energy consumption than training for the same task on a single GPU. Additionally, a higher memory consumption for single GPU training tends to lead to a higher energy consumption. This can be partially explained, as up to 30\% of the power consumption of a GPU comes from its memory for server setups \cite{gpumem2}.

In the medical imaging domain, the application of AI models to complex datasets such as MRIs, CT scans, and electron microscopy images has gained significant attention. This is mainly due to the complex nature of medical data and the substantial human resources required to analyze them and draw conclusions \cite{bioengineering10121435,Tang2020}. Among these applications of AI models in the medical field, image generation has emerged as a particularly prominent area of focus \cite{diffusionmedical}. The collection of medical imaging data is often challenging and expensive, leading to smaller datasets compared to nonmedical datasets \cite{diffusionmedical,landau2019datasetgrowthmedicalimage}. Consequently, image generation models are increasingly utilized for data synthesis, enabling data set augmentation and providing educational value for training and practice \cite{diffusionmedical}.

However, a notable gap exists in the current landscape of medical generative models: Advances in memory efficient techniques, which have been successfully applied to segmentation models \cite{cao2021swinunetunetlikepuretransformer}, have yet to be extended to generative models. The added complexity inherent in generative models poses unique challenges, particularly for 3D data sets, where the computational and memory requirements are significantly higher. Moreover, the cost of training these large-scale generative models remains prohibitively high, underscoring the need for innovations that address memory efficiency without compromising performance \cite{shi2023deepgenerativemodels3d,UZUNOVA2020101801}.

Among the image-generative models, diffusion models have become state-of-the-art techniques for the generation of high-quality images. 
However, most of these models require excessive computational resources, in particular, a substantial amount
of memory during the training phase \cite{ulhaq2024efficientdiffusionmodelsvision}. Such models are susceptible to a heavy memory load on GPUs due to the increase in the
size of the stored activations during backpropagation \cite{etmann2020iunetsfullyinvertibleunets}. This increased memory usage poses two main challenges:
(a) More computational resources require more generation of power, and (b) more memory consumption results in more heat generation. 
Both contribute adversely to global warming. Thus, it is increasingly important to develop models that reduce computational resources needed to enable sustainable advancements in image generation \cite{douwes2021energyconsumptiondeepgenerative}.

At the same time, generative flow models (GFMs) offer a promising alternative by mitigating the memory overhead associated with training \cite{bengio2021flownetworkbasedgenerative}. These models operate by mapping simple probabilistic distributions to more complex ones through a sequence of invertible transformations, based on the principles of normalizing flows \cite{bengio2021flownetworkbasedgenerative}. The invertibility of these transformations facilitates efficient memory usage, as intermediate activations do not need to be stored explicitly for backpropagation \cite{orozco2023invertiblenetworksjljuliapackagescalable}. This characteristic endows GFMs with a substantial advantage in handling high-dimensional image data, making them a viable candidate for applications where memory efficiency is paramount. However, GFMs are generally more unstable during training \cite{liu2023diffusionmodelassistedsupervisedlearninggenerative} and tend to produce lower quality samples compared to diffusion models. 

To address these challenges, we propose the Invertible Diffusion Model (IDM), which aims to merge the memory efficiency of generative flow networks with the performance and stability of diffusion models for high-dimensional medical data. IDM leverages Invertible U-Nets \cite{etmann2020iunetsfullyinvertibleunets}, integrated with invertible attention modules \cite{sukthanker2022generativeflowsinvertibleattentions,zha2021invertibleattention},\ to create a hybrid normalizing flow U-Net architecture. In our hybrid approach, we introduce invertible mappings \( f_\phi \) such that \( x = f_\phi(z) \) and \( z = f_\phi^{-1}(x) \).
This achieves memory-efficient diffusion, thereby enabling effective single GPU training. We also extend the idea of invertible attention to three-dimensional tensors. The memory efficiency arises from the property that these mappings do not require explicit storage of intermediate states during backpropagation.

By integrating invertible U-Nets with attention mechanisms extended to 3D tensors, IDM successfully combines the strengths of both DDPMs and GFMs, paving the way for efficient and scalable image generation in high-dimensional spaces.

Moreover, due to the independence of model depth and memory usage during training, complex models that capture intricate data distributions (such as 3D medical data) become more feasible for training with limited resources \cite{etmann2020iunetsfullyinvertibleunets}. The model's ability to learn invertible upsampling and downsampling layers allows it to be much more expressive and complex compared to approaches like gradient check pointing or SOTA method, which still result in higher memory peak consumption during backpropagation.

We present IDM as a superior and more robust alternative to checkpointing. Additionally, this model is portable with data augmentation and loading techniques, enabling pipelines that lead to low memory usage during both training and inference. The current SOTA memory-efficient generative models use significantly more energy as compared to IDM.

We validate the efficacy of IDM on the BraTs2020 MRI dataset for 3D image generation. We use the T1 and T2 modalities for training. Our experiments demonstrate that IDM can handle complex image generation tasks with improved memory efficiency and stability, while reducing energy consumption and improving performance.

\section{Overview and Background}
\subsection{Invertible U-Nets.}
\label{subsec:1}
Etmann et al. ~\cite{etmann2020iunetsfullyinvertibleunets} introduced fully invertible U-Nets (iUNet) as a novel approach to enhance the memory efficiency of the conventional U-Net model \cite{oktay2018attentionunetlearninglook}. This addressed a fundamental challenge in U-Net architecture, which arises from the non-bijective nature of its upsampling and downsampling processes, the paper pioneers the concept of learnable invertible upsampling and invertible  downsampling operators to achieve invertibility. Typically, U-Nets use max pooling or other non-invertible operators for downsampling and upsampling \cite{ronneberger2015unetconvolutionalnetworksbiomedical}, respectively, but by definition, these non-linear functions are not invertible. However, by introducing learnable upsampling and downsampling operators, ~\cite{etmann2020iunetsfullyinvertibleunets} was able to preserve channel expansion and reduction typical in U-Net architectures while adding constraints to allow for the channels expansion operators to be inverted. Invertible operators enable memory-efficiency by allowing for activations that normally are stored due to inability to re-calculate them to be accurately re-calculated by undergoing a full inverse pass of the model enabling a memory-efficient gradient \cite{etmann2020iunetsfullyinvertibleunets}. Building upon that, proposing and demonstrating the feasibility of orthogonal upsampling and downsampling operators as convolutions establishes the foundation for achieving invertibility. Thus, with invertible upsampling and downsampling operators, the only memory constraint of the U-Net model becomes the size of the model itself and not the additional memory that is required to store any activations associated with weights or layers.
Moreover, beyond its primary application in segmentation tasks, the versatility of iUNet is showcased through its adaptation for normalizing flow. Using downsampling operators, the paper \cite{etmann2020iunetsfullyinvertibleunets} illustrates how iUNet can allow for 2D image generation. This extension underscores the broad applicability of the iUNet architecture beyond its original scope, opening doors to diverse domains of image processing and generative modeling due to the complex representations it can capture.

\subsection{Invertible Attention.}
\label{subsec:2}
Self-Attention mechanisms have become a popular way to learn long-range data dependencies \cite{vaswani2023attentionneed,oktay2018attentionunetlearninglook,Brauwers_2023}. In the scope of images, traditional Convolutional Neural Networks lack the ability to learn context features throughout the data due to their local receptive fields. Visual transformers compute the relationships between different locations in the feature map allowing for longer range data dependencies to be learned \cite{dosovitskiy2021imageworth16x16words}. Invertible attention is a subset of attention that enables inverse operations to reconstruct the input, given the output from the attention module. 

This idea was first proposed by Zha\cite{zha2021invertibleattention}. It couples invertible residual blocks with attention modules that use Gaussian, embedded Gaussian, dot-product or concatenation-based attention functions ($F$) produces some $Q$, $K$, or $V$ and is invertible) that enable invertibility with constrained Lipschitz constant. We initially used this for our invertible attention modules but found it led to high computational cost and did not scale well to 3D data. 

Another invertible attention module was proposed by Sukthanker\cite{sukthanker2022generativeflowsinvertibleattentions}. While the focus of the paper was applying their fully-invertible attention modules with generative flow networks, their ideas extend beyond it. In particular, they use a spatial checkerboard pattern for the attention input to split the input and then process one of the inputs with their attention weights. They propose two flavors of these invertible attention modules, iMap and iTrans. iMap exploits map based attention \cite{park2018bambottleneckattentionmodule,wang2020attentionnasspatiotemporalattentioncell,woo2018cbamconvolutionalblockattention} to define a sequence of invertible functions that scale the feature map using the learned attention weights, which encode the importance of individual dimensions along the attention dimension. iTrans uses invertible convolutions \cite{kingma2018glowgenerativeflowinvertible} to obtain the key, query, and uses the input feature maps to get the value, as is standard with transformers. Using the spatial checkerboard pattern from before, enables invertibilty. To extend this work to the 3D space, we modify the iTrans model to include a depth dimension. This allows patterns based upon 3D spaces to be learned rather than taking multiple 2D slices of 3D images and learning localized patterns.

\subsection{Diffusion.}
\label{subsec:3}
Diffusion models, initially introduced by Sohl-Dickstein et al. \cite{sohldickstein2015deepunsupervisedlearningusing}, have emerged as powerful generative models for a variety of tasks and modalities \cite{yang2024diffusionmodelscomprehensivesurvey}. These models often utilize U-Net architectures as denoising agents, enabling the generation of images from randomly sampled noise based on a given timestamp \cite{chen2024overviewdiffusionmodelsapplications}. Thus, the design of the U-Net critical to the success of diffusion models \cite{williams2024unifiedframeworkunetdesign}, and significant advancements have been made to enhance the denoising process and improve image generation quality \cite{yang2024diffusionmodelscomprehensivesurvey}.

Ho et al. further refined the modeling process for diffusion models, demonstrating improved performance across various datasets \cite{ho2020denoisingdiffusionprobabilisticmodels}. These developments in 2D diffusion models led to their application in generating 3D models. Such approaches have achieved great results, particularly in generating high-quality 3D objects from text prompts \cite{lin2023magic3dhighresolutiontextto3dcontent,liu2023zero1to3zeroshotimage3d,poole2022dreamfusiontextto3dusing2d}.

Diffusion models have also been applied in the medical domain \cite{diffusionmedical} for tasks such as segmentation \cite{Fernandez_2022,kim2023diffusionadversarialrepresentationlearning,wolleb2021diffusionmodelsimplicitimage} and image generation \cite{ali2022spotfakelungsgenerating,pinaya2022brainimaginggenerationlatent,takezaki2024ordinaldiffusionmodelgenerating,yu2024ctsynthesisconditionaldiffusion}. Given the inherently 3D nature of many medical datasets, researchers have extended diffusion-based models to 3D applications in the medical space, targeting tasks like generating and segmenting medical imagery for conditions such as brain tumors, pulmonary nodules, liver tumors, Alzheimer’s disease etc. \cite{khader2023medicaldiffusiondenoisingdiffusion,Wang_2022}.

Despite these successes, modeling medical imagery, such as MRI and CT scans, poses unique challenges due to their high complexity and the substantial memory and processing requirements \cite{guo2024maisimedicalaisynthetic,UZUNOVA2020101801}. Training these 3D diffusion models often demands multiple GPUs, leading to increased energy consumption and memory usage during both training and inference.

\section{Related Works}
For memory-efficient training of Diffusion models there have been a few novel approaches that tackle the issue. The most relevant paper was proposed by Bieder et al. \cite{bieder2023diffusionmodelsmemoryefficientprocessing} for memory-efficient diffusion model training of 3D medical imagery. The PatchDDM model presented performs training on random patches of 3D volumes for training but uses the entire volume during inference leading to significant memory savings. This approach enables smaller models to be trained on smaller sizes of data, allowing for memory-savings during training as well. In this space of PatchDDM outperforms other models both in terms of performance as well as memory-efficiency. On the other hand, IDM’s architecture is fundamentally designed to handle entire volumes during training, thereby avoiding any implicit biases about the data and its properties. Although the data augmentation techniques employed by PatchDDM could potentially be integrated into IDM, to the best of our knowledge, doing so would compromise IDM’s versatility and generalizability across both 2D and 3D data. Work in diffusion models for medical images space require large GPU memory consumption during training due to a lack of memory-specific data augmentation or model changes to reflect the more complex and large data \cite{Montoya_del_Angel_2024}.
Another memory-efficient patch-based model was proposed by Arakawa \cite{arakawa2023memoryefficientdiffusionprobabilistic}. This differed however, as its focus was for 2D datasets and memory consumption during inference by partitioning input data into patches to reduce input and output sizes. Thus, images could be generated patch-by-patch rather than all at once. The ideas presented could easily be implemented along with IDM for a reduction in memory usage during inference.
In the realm of medical image synthesis, there exist many models for generative 3D images for T1 and T2. To the best of our knowledge, the current SOTA is AGGrGAN \cite{mukherkjee2022brain} which aggregates three base GAN models—two variants of Deep Convolutional Generative Adversarial Network (DCGAN) and a Wasserstein GAN (WGAN) to generate synthetic MRI scans of brain tumors. There also exists a CycleGAN approach that exploits the ability to get 2D slices from 3D data \cite{cyclegan}. Another popular synthesis method for MRI data uses cross-modality models (T1 to T2 and T2 to T1 for example) with MT-Net \cite{mtnet} being a very well performing model that exploits the multi-modal nature of MRI data for synthesis using transformers.

\section{Invertible Diffusion Model}
\label{sec:4}

For the IDM, we take the invertible UNet idea presented by Etmann et al. \cite{etmann2020iunetsfullyinvertibleunets} and couple it with the invertible attention ideas introduced in \cite{sukthanker2022generativeflowsinvertibleattentions,zha2021invertibleattention}. This allows us to reconstruct activations rather than storing them, as is standard with diffusion models, during backpropagation at the cost of an increase in computations required. This cost is justified however, as models limited by computational power still are able to undergo training as opposed to models limited by VRAM memory, which are very likely to have out of memory errors during training. This leads to longer training times for the model. Additionally, even though there is an increase in the number of computations done, the relative energy consumption will not scale linearly with it because we are using and managing less memory. Also, due to iUNets learnable downsampling and upsampling, the memory demand is (mostly) independent of the depth of the network \cite{etmann2020iunetsfullyinvertibleunets}. This allows for deeper networks without the added cost of an increase in memory consumption during backpropagation. So, IDM enables deeper, more energy and memory efficient networks.
During channel concatenation between the upsampled channels and the split channels from the downsampling process \cite{etmann2020iunetsfullyinvertibleunets}, the split channels go through an invertible attention module. This enables split channels during downsampling to be reconstructed during backpropagation. We use the invertible transformer attention module (iTrans) from \cite{sukthanker2022generativeflowsinvertibleattentions}. During the forward propagation, the channel concatenation operations are as follows:

\begin{equation}
 a = y_2 \odot f(y_1)
 \label{1}
\end{equation}
\begin{equation}
 y^U_i = concat(a,c) 
\label{2}
\end{equation}

Where $f(.)$ in (1) is the attention weight computation for iTrans and $\odot$ is element-wise matrix multiplication. Using the checkerboard masking presented in \cite{sukthanker2022generativeflowsinvertibleattentions}, we get $y_1$ and $y_2$ from $y$ where $y$ are the split channels from downsampling layer $i$.  $ y^U_i $ is the concatenated image from the downsampling layers and the current upsampling layer, both at level $i$. 
Due to both iUNets' and iTrans' invertibilty we can define an invertible function to reconstruct the split channels from the downsampling layer ($y$):
\begin{equation}
y_1,y_2 =split(y^U_i)
\label{3}
\end{equation}
\begin{equation}
 y_2 = a  \varnothing f(y_1)
\label{4}
\end{equation}
where $\varnothing$ is element wise division. All equations are taken and modified from \cite{etmann2020iunetsfullyinvertibleunets} and \cite{sukthanker2022generativeflowsinvertibleattentions}. These equations describe the forward and backward process for the attention module, enabling invertibility. Since we can reconstruct these split channels, we are able to save memory during training by being able to reconstruct activations instead of storing them for backpropagation for all layers in the model, including the attention networks for each layer. Sukthanker et al. applied these ideas only to 2D images, we extend these to 3D images. In total, the Diffusion process is described by:
\begin{equation}
q(x_t | x_{t-1}) = \mathcal{N}(x_t | \sqrt{1-\beta_t}x_{t-1}, \beta_t I)
\label{5}
\end{equation}
\begin{equation}
q(x_t | x_{t-1}) = \mathcal{N}(x_t |\sqrt{\Bar{a}_t} x_0, (1-\Bar{a}_t)I)
\label{6}
\end{equation}
\begin{equation}
x_{t-1} = \frac{1}{\sqrt{a_t}} \left(x_t - \frac{\beta_t}{\sqrt{1-\bar{\alpha}_t}} U(x_t,t)\right) + \sqrt{\beta_t}\epsilon
\label{7}
\end{equation}
Where $t$ is a timestamp, $\beta_t$ is the variance at timestep $t$, and $x_t$ is approximately sampled from $ \mathcal{N}(0,1)$ for Equation \ref{5}. In Equation \ref{6}, $\bar{a_t} = 1 - \beta_t$ and 
$\bar{\alpha}_t = \prod_{s=1}^t \alpha_s$. Equations are taken and modified from \cite{arakawa2023memoryefficientdiffusionprobabilistic,ho2020denoisingdiffusionprobabilisticmodels}. Equations \ref{5} and \ref{6} define how our latent representation can be modified to look like Gaussian noise \cite{arakawa2023memoryefficientdiffusionprobabilistic,ho2020denoisingdiffusionprobabilisticmodels}. Equation \ref{7} defines the reverse process for de-noising our Gaussian noise for image generation. $U(\cdot)$ is our UNet backbone that allows for memory-efficient training, as is standard with Diffusion models \cite{calvoordonez2024missinguefficientdiffusion,ho2020denoisingdiffusionprobabilisticmodels}, and $\epsilon \sim \mathcal{N}(0,1)$.

For our loss we use a combination of mean squared error losses for the forward and inverse pass to align both as well as L2 regularization.
\begin{equation}
\begin{aligned}
\mathcal{L} = & \mathbb{E}_{t,\mathbf{x}_t,\epsilon} \left[ \left\| \epsilon - U(x_t, t)\right\|^2 \right] \\
               &+ \lambda_{\text{R}} \mathbb{E}_{t,\mathbf{x}_t} \left[ \left\| x_t - U^{-1}(U(x_t,t))\right\|^2 \right] \\
               &+ \lambda_{\text{L2}} \left\| W \right\|_2,\end{aligned}
\end{equation}
where, $U(\cdot)$ is the forward pass of our model and $U^{-1}(\cdot)$ is the inverse pass. $\epsilon$ is a sample from a Gaussian distribution and $t$ is the timestamp. It is important to note how $U^{-1}$ does not take in a timestamp. We conjecture, doing so would introduce memory consumption issues that would propagate throughout the whole network. However, our experiments have shown that this still leads to stable training. $\lambda_{\text{R}}$ is a hyperparameter between 0 and 1 to tune how important the forward and inverse pass is for the loss. From our experimentation, higher values of $\lambda_{\text{R}}$ means our approach will be able to reconstruct our inputs better given our output. In doing so, our forward and inverse passes grow more aligned which could be useful depending on the task, at the cost of increased FLOPs. We also found that introducing weight regularization with $W$ as the model's weights and $\lambda_{\text{L2}}$ led to better performance.

We make a modification on top of the iUNets architecture to allow for complex data distributions to be learned. iUnets' learnable invertible downsampling layers require that spatial dimensions be divisible by the stride of the downsampling layer \cite{etmann2020iunetsfullyinvertibleunets}. Since most image data contains 1 to 3 channels, this leads to very weak models as layers can only double or triple the number of channels in previous layers. As is suggested by \cite{etmann2020iunetsfullyinvertibleunets} and from our own experiments, to remedy these issues, we introduce non-invertible non-linearly increasing downsample and upsample layers at the start and end of the model, respectively. In practice, we found it was best to leave these layers as non-invertible and instead store their activations, leading to a slight increase in memory consumption during training for much greater model power. We found that not including these additional layers led to the model being unable to generate any images and collapse quickly during training. A visualization of the UNet model architecture can be seen in Figure 1.
\begin{figure}
    \centering
    \includegraphics[width=0.99\linewidth]{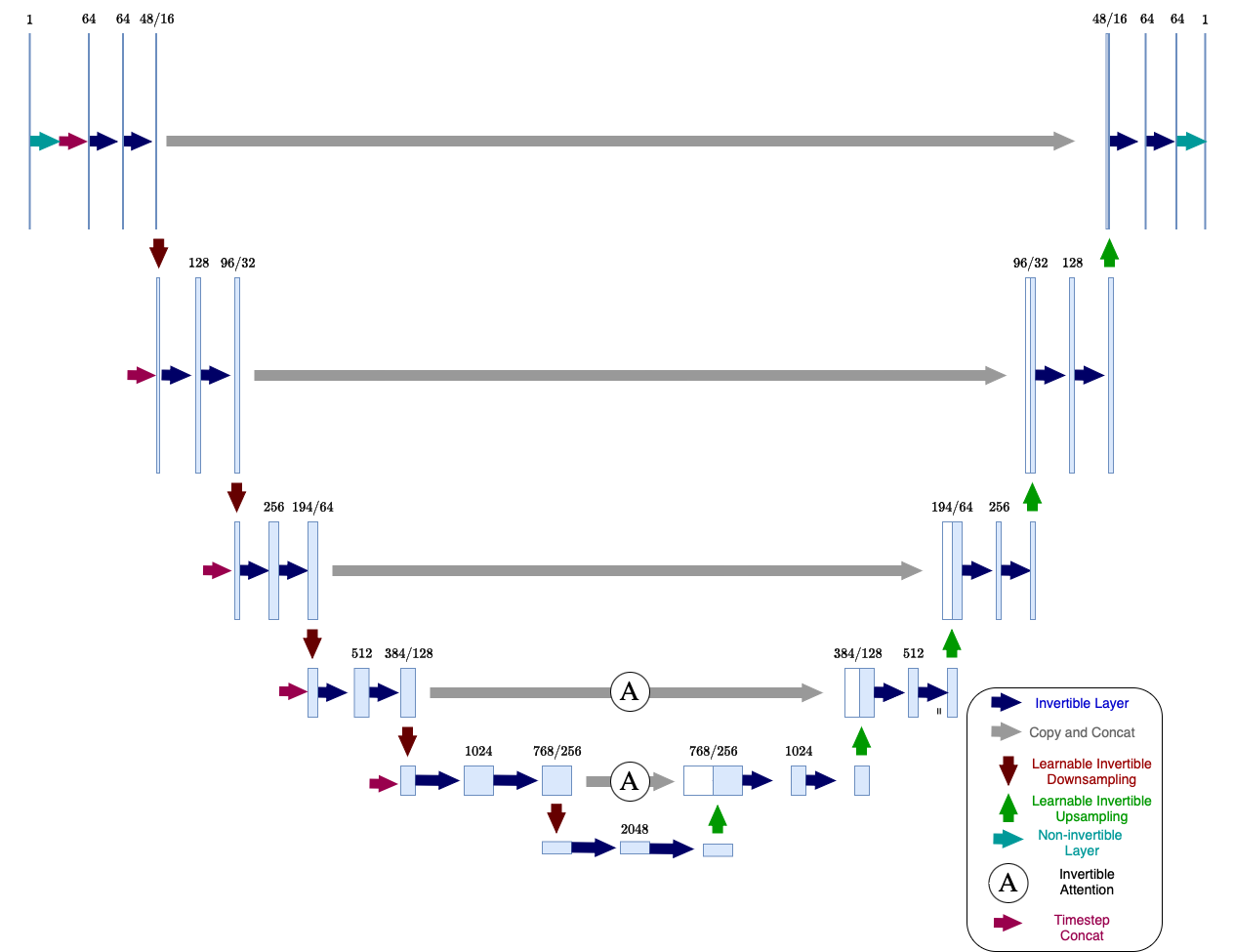}
    \caption{Modified iUNet with invertible attention modules }
    \label{fig:diag}
\end{figure}
\section{Experiments}
\label{sec:5}
\subsection{Datasets}
\label{subsec:1}
The BraTS2020 dataset contains 369  MRI scans  with a resolution of 1 × 1 × 1 $mm^3$ \cite{brats2,brats4,brats5,brats3,brats1}. Each scan has dimensions of 240 × 240 × 155. We mostly follow the produce for reshaping the MRI scans into 64 x 64 x 64 voxels from \cite{khader2023medicaldiffusiondenoisingdiffusion,kwon2019generation}. We also included normalization to ensure values are between [0,1]. BraTS2020 has four modalities T1, T1ce, T2, and FLAIR. For this paper, we use T1 and T2 since this simplifies the problem by focusing on structural and anatomical features, which are well represented in these sequences. T1 highlights non-enhancing tumor core and enhancing tumor, while T2 highlights edema, making them sufficient for analyzing two important types of tumors \cite{brats1}. Additionally, T1 and T2 are typically available across datasets, which helps ensuring broader applicability in real-world clinical scenarios.

\subsection{Experiment Details}
\label{subsec:2}
For the BraTS2020 dataset, due to the large voxel sizes, we are using a single NVIDIA RTX A5000 with 24 GB of VRAM. This setup ensures the efficient processing of high-resolution 3D volumes typically used in medical imaging tasks. The model architecture has the channel sizes for the encoder and decoders portions as: (64, 128, 256, 512, 1024, 2048) to capture fine-grained details and support deeper representations. We include the invertible attention module in the last two layers. The model is trained using the DDPM \cite{ho2020denoisingdiffusionprobabilisticmodels} approach with a noise scheduler utilizing a cosine beta schedule, with 2000 time steps. Training is with a learning rate of 2e-4 and AdamW optimizer. A batch size of 4 is used to accommodate the large voxel size. The training environment is managed by PyTorch with the use of a cosine annealing learning rate scheduler to adjust the learning rate over time.  To further improve generalization, a slight L2 regularization by setting $\lambda_{\text{L2}}$ to 0.0001 for the loss function. Since this is not a reconstruction task as the U-Net is predicting noise rather than reconstructing the input, we set  $\lambda_{\text{R}}$ to 0.

\subsection{Evaluation Details}
\label{subsec:2}
The model's performance is evaluated using Peak Signal-to-Noise Ratio (PSNR) Structural Similarity Index (SSIM) and Mean Absolute Error (MAE), providing both noise prediction accuracy and image reconstruction quality metrics. We report the best results on a batch, with the batch size of 2, of the validation dataset. We also determine the FLOPs the model performs during one forward and backward pass and the peak memory overhead associated with it. Additionally, we monitor the temperature and energy usage during the first two minutes of forward and backward passes.

\section{Results}
The performance of IDM was evaluated on the BraTs2020 dataset for T1 and T2 modalities and compared against SOTA models. The results highlight IDM's ability to balance memory efficiency, computational complexity, and energy usage while synthesizing images. We also evaluated IDM on SynapseMNIST3D \cite{medmnistv2}, which contains electron microscopy data, and achieved reasonable results but due to space constraints and BraTs being significantly more popular and clinically relevant, choose to not include those results.

\label{sec:6}

\begin{table}[hp]
\caption{Comparison of PSNR, SSIM, and MAE for Different BraTs2020 T1 and T2 modalities. The best result on a single batch from the validation set for IDM shown. For other models, we take results from their respective papers.}

\centering
\begin{tabular}{@{}lccc@{}}
\toprule
\textbf{T1}             & \textbf{PSNR} & \textbf{SSIM} & \textbf{MAE} \\ \midrule
IDM \textbf{(Ours)}                    & 25.01         & 0.4852        & 0.4090        \\
AGGrGAN                  & 22.12         & 0.77          & -            \\
CycleGANs      & 34.244        & 0.983         & -         \\
MT-Net                  & 22.193        & 0.906         & -            \\ \midrule
\textbf{T2}             & \textbf{PSNR} & \textbf{SSIM} & \textbf{MAE} \\ \midrule
IDM \textbf{(Ours)}                     & 25.54         & 0.4879        & 0.3349       \\
AGGrGAN                 & 21.94         & 0.8           & -            \\
CycleGANs      & -             & -             & -            \\
MT-Net                  & 22.028        & 0.903         & -            \\ \bottomrule
\end{tabular}
\label{tab:performance}
\end{table}

\subsection{Quantitative and Qualitative Results}
\begin{figure} [htb]
    \centering
    \includegraphics[width=.85\linewidth]{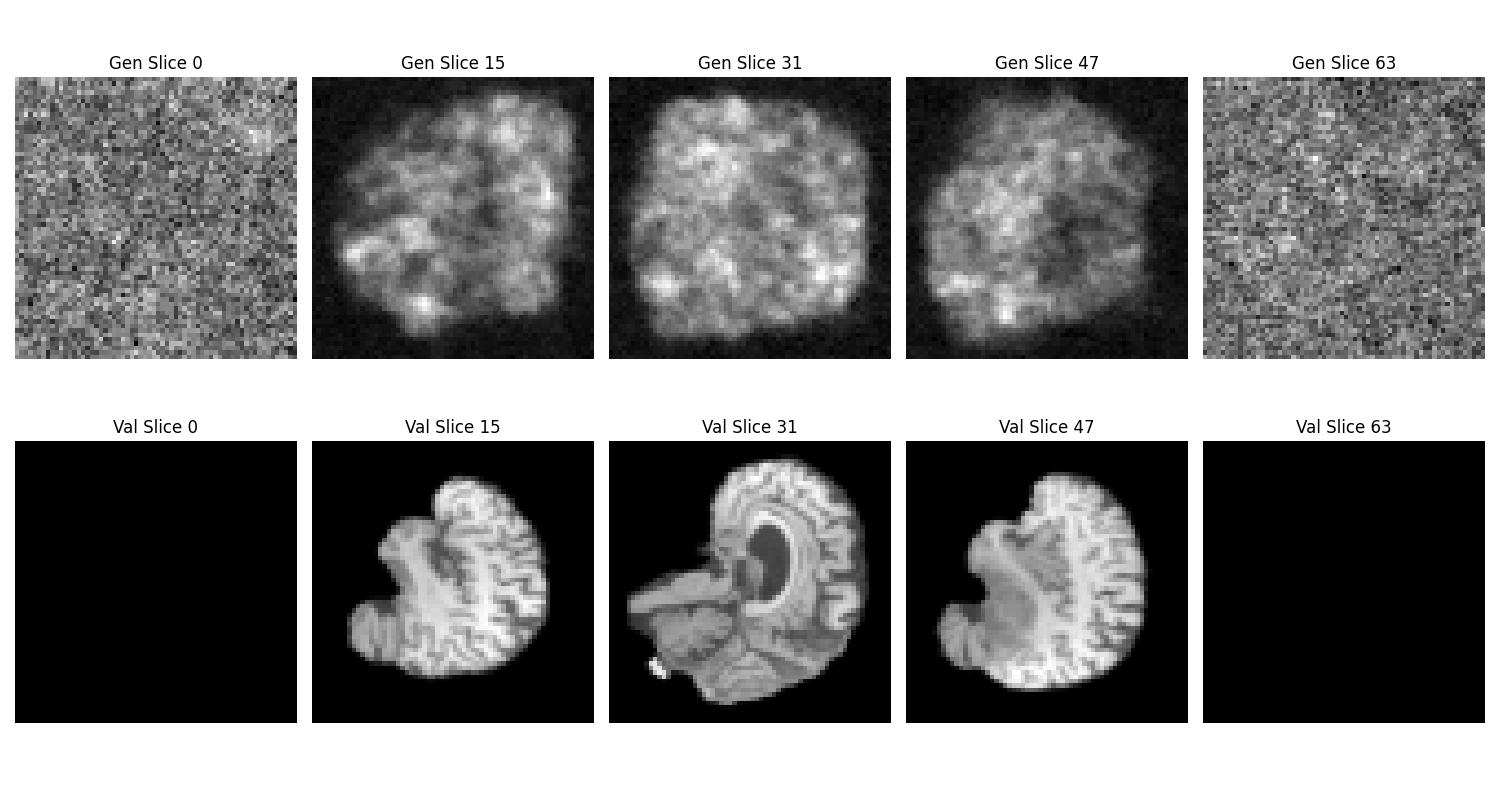}
    \caption{Generated MRI Slice compared to Real MRI Slice}
    \label{fig:gen}
\end{figure}

For both T1 and T2 modalities, it can be seen that IDM performs as well as most other models in terms of PSNR but has a significantly lower SSIM. This discrepancy might arise from IDM’s reliance on a simple U-Net architecture, minimally modified from the diffusion model in \cite{ho2020denoisingdiffusionprobabilisticmodels} and the invertible U-Net in \cite{etmann2020iunetsfullyinvertibleunets}. The generated slices in Figure~\ref{fig:gen} show good shape reconstruction, aligning with real MRI slices. This alignment is reflected in the high PSNR scores in Table \ref{tab:performance}. However, the lower SSIM scores indicate a lack of fine-grained structural fidelity, which can be observed in the lack of detail and structure in the generated images compared to the real ones. While this trade-off may limit IDM’s utility in medical imaging applications requiring high structural fidelity, its performance suggests potential applicability in other downstream tasks where pixel-wise accuracy is prioritized.

\begin{figure}[htbp]
    \centering
    \includegraphics[width=0.5\textwidth]{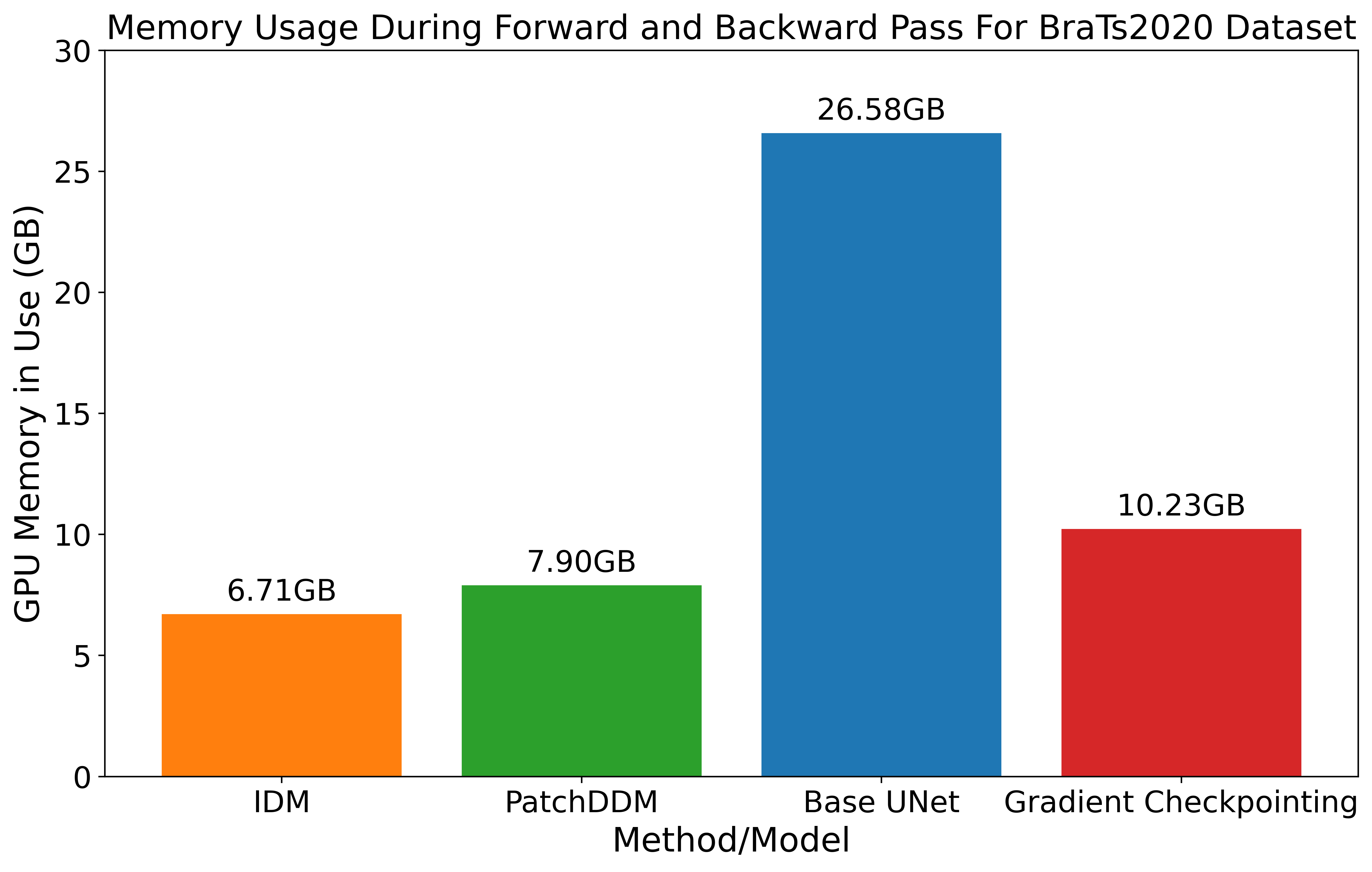}
    \caption{Maximum memory allocated on GPU during the training phase for various methods/models}
    \label{fig:gpumems}
\end{figure}

\subsection{Efficiency and Resource Utilization}
\begin{table*}[htbp]
\centering
\caption{Comparison of FLOPs, peak memory consumption, and peak temperature for one training step between PatchDDM-3D and IDM. Ratios represent IDM relative to PatchDDM-3D.}
\label{tbl:runtime_results}
\begin{tabular}{lcc}
\textbf{Metric}                   & \textbf{PatchDDM-3D} & \textbf{IDM (Relative Ratio)} \\ 
\midrule
FLOPs (Count)                     & 1,225,184,889        & 5,643,083,550 (\textbf{4.61×}) \\ 
Peak Memory Consumption (GB)      & 7.899565             & 6.708984 (\textbf{0.85×}) \\ 
Peak Temperature (°C)             & 62                   & 61 (\textbf{0.98×}) \\ 
\bottomrule
\end{tabular}
\end{table*}

IDM demonstrates notable trade-offs between memory efficiency, computational complexity, and energy consumption. Based on table \ref{tbl:runtime_results} It achieves a 15\% reduction in peak memory usage during the forwards and backwards pass compared to PatchDDM-3D, despite having approximately 14× more parameters (343.6M vs. 24.3M). These memory savings can be attributed to the invertible nature of IDM. This feature allows IDM to train larger models in memory-constrained environments (e.g. single GPU training) by allocating more space on the GPU for the data and model itself and not the activations. This is further seen in Figure \ref{fig:gpumems} where IDM requires the least amount of memory compared to SOTA models and popular memory reduction methods.

However, this greater memory efficiency comes at the cost of higher FLOPs per training step compared to PatchDDM-3D (4.61x higher). This is due to the inverse computations required by IDM during back propagation and the larger parameter count, which increases the number of operations per forward and backward pass. Despite this, IDM demonstrates efficient parameter utilization as although it has 14x more parameters, it performs only 4.61x more FLOPs, showcasing a favorable trade-off between model size and computational efficiency.

\begin{figure}[htbp]
    \centering
    \begin{minipage}{0.49\textwidth}
        \centering
        \includegraphics[width=\linewidth]{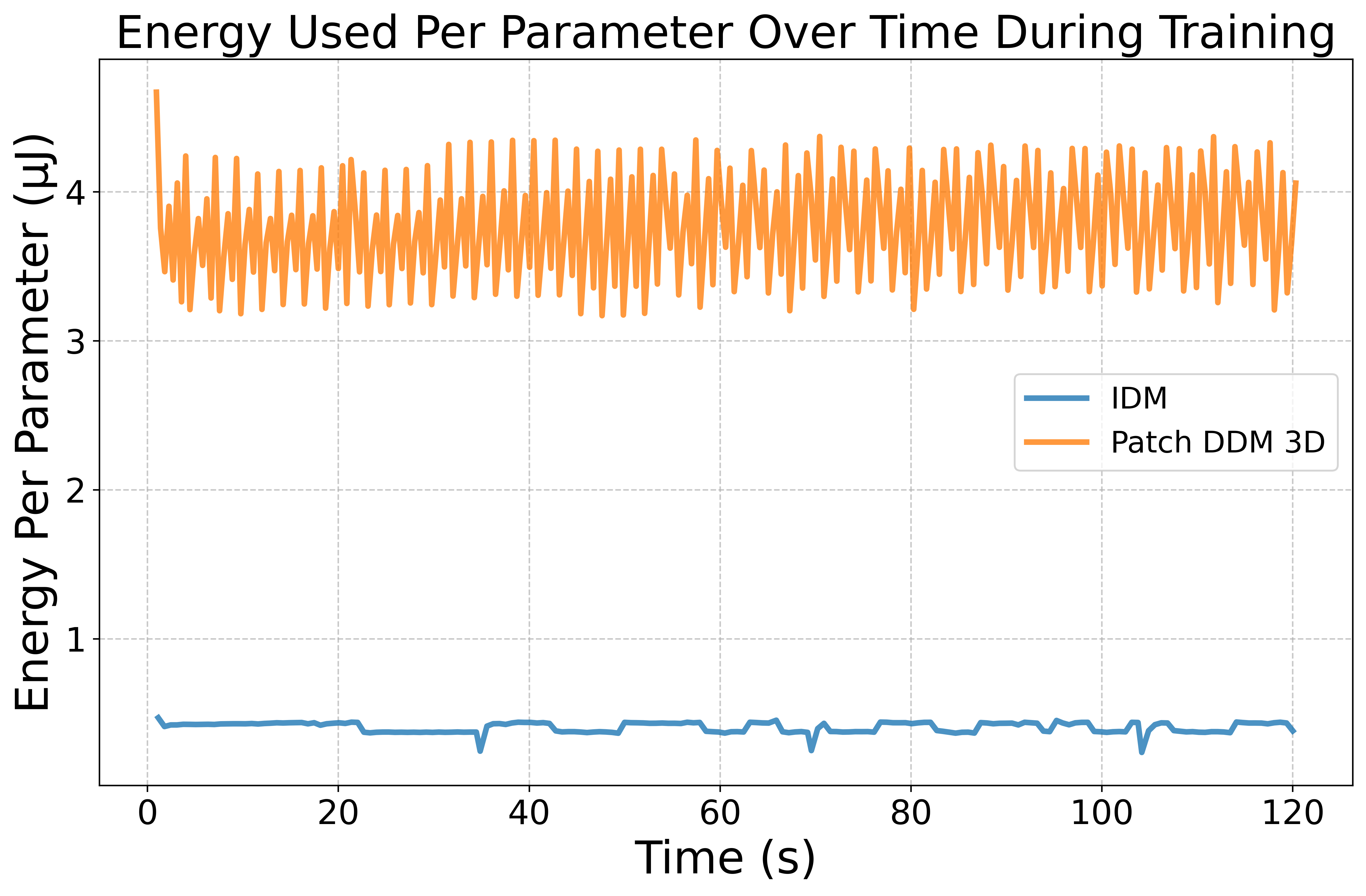}
        \caption{Energy used per parameter over the first two minutes of forward and backward passes}
        \label{fig:energy_over_time.png}
    \end{minipage} \hfill
    \begin{minipage}{0.49\textwidth}
        \centering
        \includegraphics[width=\linewidth]{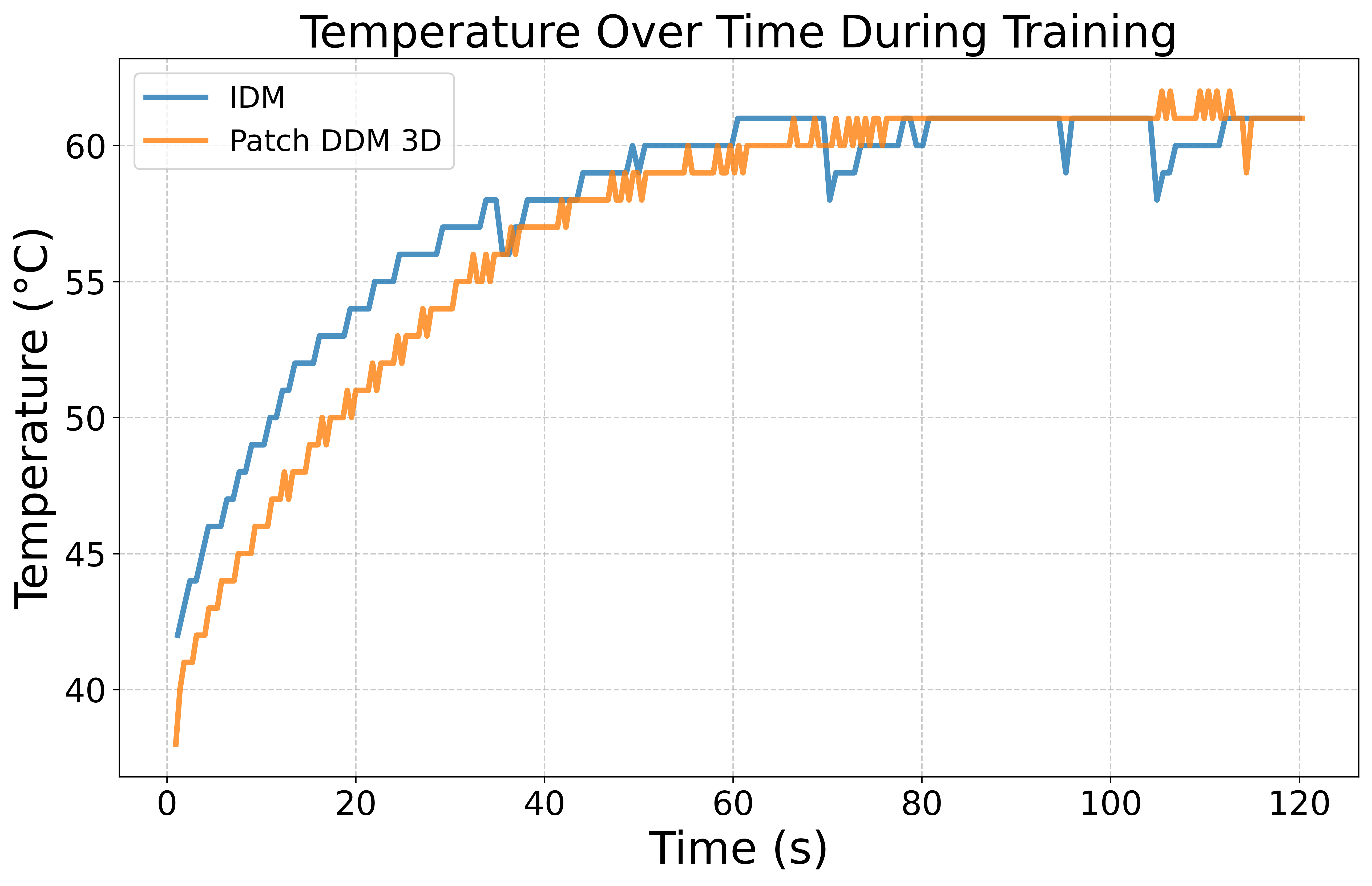}
        \caption{Temperature used over the first two minutes of forward and backward passes}
        \label{fig:temperature_over_time}
    \end{minipage}
\end{figure}

Energy efficiency is another notable strength of IDM. Despite the increased FLOPs, IDM consumes significantly less energy per parameter over time compared to PatchDDM-3D as seen in Figure \ref{fig:energy_over_time.png}. This is due to the invertible architecture leading to less memory needing to be managed and effectively using its parameters. This energy efficiency also underscores IDM’s scalability as even with very large models (343.6M parameters), it uses energy effectively. It is key to note that IDM initially generates slightly higher GPU temperatures compared to PatchDDM-3D but stabilizes over time based on Figure \ref{fig:temperature_over_time}. However, this does not discount the possibility that for even larger models with greater model capacity and better SSIM results, the temperature could rise further, potentially offsetting some of the energy and memory efficiency gains observed in IDM currently.

\section{Conclusion}
We introduce the invertible diffusion model (IDM) which enables significantly memory-efficient training for diffusion modules by removing the need to store activations for backpropagation, and instead re-computing them through inverse operations by modifying the UNet backbone \cite{etmann2020iunetsfullyinvertibleunets}. This has the bonus effect of reducing energy-consumption for the model as well. Our experiments show reduction in memory consumption during training. It scales particularly well in 3D medical imagery datasets. Due to the increase in dimensionality of such datasets, they typically consume more memory during back propagation. This is due to their networks being deeper and larger which leads to more activations and these activations require more space to store. We hope this work will allow for complex diffusion models with high-dimensional data to be more easily and sustainably trained with a decrease in the GPU resources being used.
\section{Future Works}
In future works, we hope to take ideas from \cite{wang2023patchdiffusionfasterdataefficient} to reduce the training and inference time for IDM by reducing the computational power needed. To extend our work, we hope to exploit the inverse prediction capabilities of IDM to predict the noise that a random image came from, allowing for noise distributions to be learned and thus enabling the creation of a classifier to distinguish between real and diffusion-generated images. Additionally, we hope to apply our model to synthesize other higher dimensional medical images.
%
%
%
%

\end{document}